Research Article

# Prognosticating Autism Spectrum Disorder Using Artificial Neural Network: Levenberg-Marquardt Algorithm

**Avishek Choudhury*, Christopher M Greene**

Binghamton University, New York, USA

***Corresponding Author:** Avishek Choudhury, PhD Student, Binghamton University, Systems Science and Industrial Engineering, Engineering Building, L2, Vestal, NY 13902, USA, Tel: +1 (806) 500-8025; E-mail: achoud13@binghamton.edu

## Abstract

Autism spectrum condition (ASC) or autism spectrum disorder (ASD) is primarily identified with the help of behavioral indications encompassing social, sensory and motor characteristics. Although categorized, recurring motor actions are measured during diagnosis, quantifiable measures that ascertain kinematic physiognomies in the movement configurations of autistic persons are not adequately studied, hindering the advances in understanding the etiology of motor mutilation. Subject aspects such as behavioral characters that influences ASD need further exploration. Presently, limited autism datasets concomitant with screening ASD are available, and a majority of them are genetic. Hence, in this study, we used a dataset related to autism screening enveloping ten behavioral and ten personal attributes that have been effective in diagnosing ASD cases from controls in behavior science. ASD diagnosis is time exhaustive and uneconomical. The burgeoning ASD cases worldwide mandate a need for the fast and economical screening tool. Our study aimed to implement an artificial neural network with the Levenberg-Marquardt algorithm to detect ASD and examine its predictive accuracy. Consecutively, develop a clinical decision support system for early ASD identification.

## 1. Introduction

Autism spectrum disorder (ASD) is a faction of polygenetic evolving brain disarray accompanied with behavioral and cognitive mutilation [1]. It is a lifelong neurodevelopmental illness depicted by insufficiencies in communication, interaction and constrained behavior [2]. Though ASD is identified mainly by behavioral and social physiognomies, autistic individuals often exhibit tainted motor ability such as reduced physical synchronization, unstable body balance and unusual posture and movement patterns [3-5]. Individuals with ASD show stereotypical recurring actions, constrained interests, a privation of instinct control, speech insufficiencies, compromised intellect and social skills compared to typically developing (TD) children [6]. There has been well-established work done in diagnosing ASD using kinematic physiognomies.

Guha et al. [7] used gesticulation data to measure atypicality variance in facial expressions of children with and without ASD. Six facial reactions were securitized using information theory, statistical analysis and time- series modeling. Recently researchers have engrossed in refining and implementing data analytics as a diagnosis tool for ASD. Being an efficient computational tool machine learning has disclosed its potential for classification in the various domain [8-9]. Therefore, some literature has employed machine learning methods to identify neural [10] [11] or behavioral markers [12-13] responsible for discriminating individuals with and without ASD. Stahl et al. [10] studied the influence of eye gaze to classify infant groups with a high or low risk of getting ASD. Computational methods such as discriminant functions analysis with an accuracy of 0.61, support vector machine with an accuracy of 0.64, and linear discrimination analysis with an accuracy of 0.56 were implemented. Bone in 2016 [14] retrieved data from The Autism Diagnostic Interview-Revised and The Social Responsiveness Scale and applied SVM machine learning classifier to a significant group individual with and without ASD and obtained promising results and demonstrated that machine learning could be used as a useful tool for ASD diagnosis. Grossi et al. employed artificial neural networks (ANNs) to develop a predictive model using a dataset comprising of 27 potential pregnancy risk elements in autism development [15]. The artificial neural network produced a predictive accuracy of 80%. Their study supported and encouraged the use of ANNs as an excellent diagnostic screening tool for ASD. The highest classification accuracy obtained so far is 96.7% using SVM.

With the aim of developing a clinical decision support system, we implement artificial neural networks with the Levenberg-Marquardt algorithm on data set that contains ten behavioral and ten personal attributes of adults with and without ASD.

## 2. Methodology

This study does not involve any participation of human subjects. We extracted the data from the UCI library. The data and data description is provided with this paper. It consists of 20 predictors (ten behavioral and ten personal attributes), one response variable ad 704 instances. The methodology designed for this study can be divided into (a) data preprocessing, (b) designing the model, and (c) fitting and evaluating the model.

### 2.1 Data preprocessing

Data preprocessing is one of the most critical steps in all machine learning application. In this study, we did not use missing data points and partitioned the dataset into training, testing and selection instances. The following pie chart (Figure 1a) details the uses of all the instances in the dataset. The total number of instances is 704, that contains 424 (60.2%) training instances, 140 (19.9%) of selection instances, and 140 (19.9%) of testing instances. The following pie chart includes all the missing values.

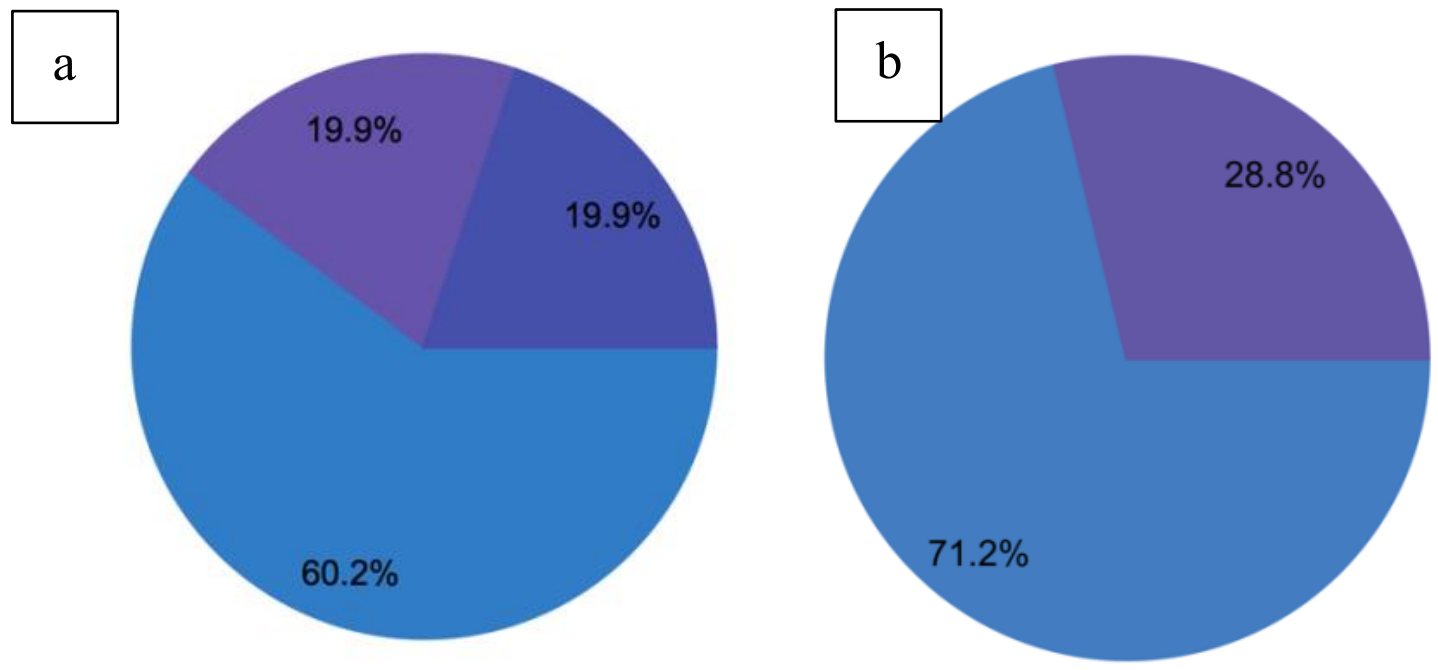

**Figure 1:** (a) Shows the partitioned data; (b) Shows the number of instances belonging to each class in the dataset.

The pie chart above (Figure 1b) displays the partitioning of the dataset (excluding all missing values). The number of instances with negative Class/ASD (blue) is 222, and the number of instances with positive Class/ASD (purple) is 90. This Figure also shows that the data is unbalanced. However, we did not implement any data balancing method.

## 2.2 Designing the model

In this step we calculated the suitable training algorithm for our dataset and determined the complexity of the model, that is the optimal number of neurons in the network.

**2.2.1 The levenberg–marquardt algorithm:** Kenneth Levenberg and Donald Marquardt developed The Levenberg–Marquardt algorithm (LM) [16-17] that generates a mathematical solution to a problem of minimizing a non- linear function. We used this algorithm because in the domain of artificial neural-networks it is fast and has stable convergence. The LM algorithm approaches the second-order training speed without calculating the Hessian matrix. It holds good when the loss function has the arrangement of a sum of squares. LM is an optimization algorithm that outperforms simple gradient descent and conjugate gradient methods in a diverse assortment of problems. LM algorithm follows equation 1 as shown below.

$$\boldsymbol{w}_{k+1} = \boldsymbol{w}_k - \left(\boldsymbol{J}_k^T \boldsymbol{J}_k + \mu \boldsymbol{I}\right)^{-1} \boldsymbol{J}_k \boldsymbol{e}_k \qquad (1)$$

where J is the Jacobian matrix, T stands for transpose, k is the index of iteration, e is the training error and w is the weight vector.

The following table 1 shows a brief description of the parameters used for this algorithm.

| Parameters | Description | Value |
|---|---|---|
| Damping parameter factor | Damping parameter increase/decrease factor. | 10 |
| Minimum parameter increment norm | Norm of the parameter increment vector at which training stops. | 0.001 |
| Minimum loss increase | Minimum loss improvement between two successive iterations | $1e^{-12}$ |
| Performance goal | Goal value for the loss | $1e^{-12}$ |
| Gradient norm goal | Goal value for the norm of the objective function gradient | 0.001 |
| Maximum selection loss increase | Maximum number of iterations at which the selection loss increases | 100 |
| Maximum iterations number | Maximum number of iterations to perform the training. | 1000 |

**Table 1:** The Levenberg-Marquardt algorithm description.

**2.2.2 Order selection algorithm:** For better performance of the model, we implemented an incremental order selection algorithm to achieve a model with the best complexity to produce an adequate fit of the data. Incremental selection order is the naivest order selection algorithm. This method begins with a least order and upsurges the size of the hidden layer of neurons until a desired order is attained. The order selection algorithm determines the optimal number of neurons in the network. Incremental order selection enhances the ability of the model to predict the result with a new data. Two recurrent problems that hinder the design of a neural network are underfitting and overfitting. The best simplification is attained by designing a model whose complexity is appropriate to produce the best fit model. Underfitting can occur if the model is too simple, whereas overfitting is defined as the effect of a selection error burgeoning due to an over-complex model. The next Figure 2 illustrates the training and selection errors as a function of the order of a neural network.

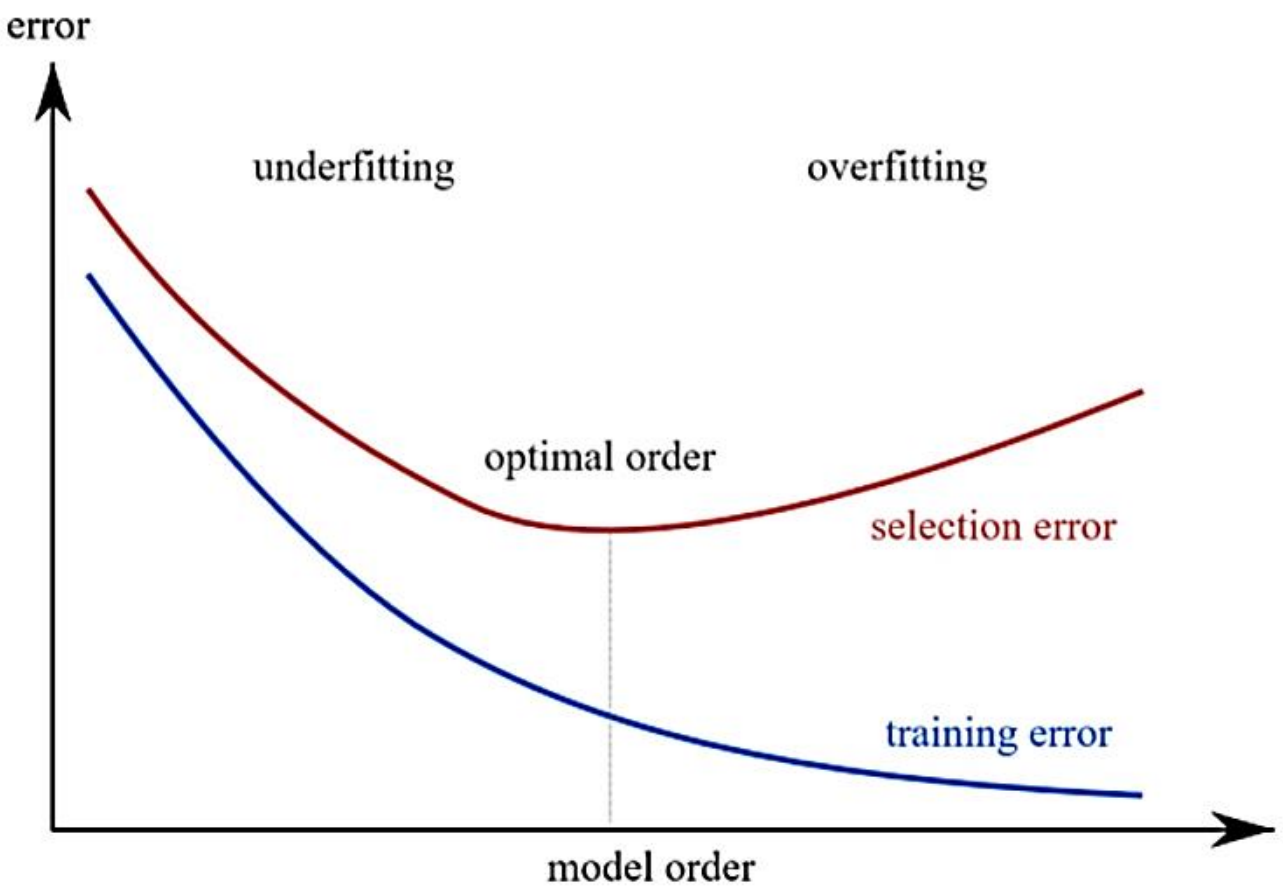

**Figure 2:** Effect of model complexity on Overfitting and Underfitting.

## 2.3 Fitting and evaluating the model

After designing the appropriate model, we deployed it on our preprocessed dataset and computed the following performance measures: (a) Accuracy, (b) Sum squared error, (c) mean squared error, (d) root mean squared error, (e) normalized squared error, (f) cross-entropy error, (g) Minkowski error, and (h) weighted squared error. Additionally, the area under the ROC curve, miss-classification, cumulative gain, lift chart and model loss index were observed for better understanding and evaluation of the model's performance.

## 3. Results

Implementing the Levenberg–Marquardt algorithm and incremental order selection enabled the proposed model to produce a classification accuracy of 98.38%. The following table 2 shows the training results obtained by the Levenberg-Marquardt algorithm. They include final states from the neural network, the loss function, and the training algorithm. The number of iterations needed to converge is found to be zero, which infers that the training algorithm did not modify the state of the neural network.

| Parameters | Value |
|---|---|
| Final parameters norm | 14.5 |
| Final loss | $9.89e^{-05}$ |
| Final selection loss | 0.00122 |
| Final gradient norm | 0.00036 |
| Iteration number | 0 |
| Elapsed time | 0 |
| Stopping criterion | Gradient norm goal |

**Table 2:** The result obtained from the Levenberg-Marquardt algorithm.

The optimal number of neurons was calculated to be 1. The Figure 3 displays the loss history for the different subsets during the incremental order selection process. The blue line represents the training loss, and the red line symbolizes the selection loss.

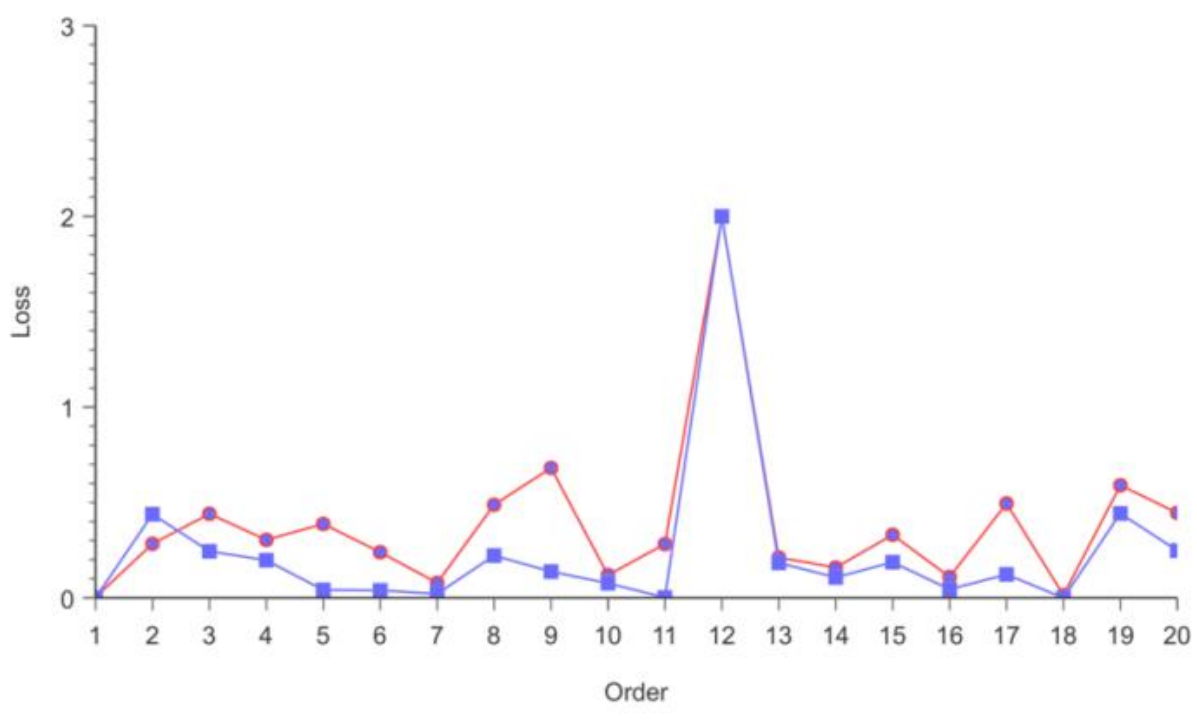

**Figure 3:** Loss history for different subsets obtained during incremental order selection algorithm.

| Parameters | Value |
|---|---|
| Optimal order | 1 |
| Optimum training loss | $8.2344e^{-05}$ |
| Optimum selection loss | 0.00313 |
| Iteration number | 20 |

**Table 3:** Final output obtained from incremental order selection algorithm.

### 3.1 Testing errors

This task measures all the losses of the model. It considers every used instance and evaluates the model for each use. The table 4 below shows all the training, testing and selection errors of the data.

| Errors | Training | Selection | Testing |
|---|---|---|---|
| Sum squared error | 0.0032 | 0.0796 | 0.3243 |
| Mean squared error | $1.7360e^{-05}$ | 0.0012 | 0.0052 |
| Root mean squared error | 0.0041 | 0.0358 | 0.0723 |
| Normalized squared error | $7.0236e^{-05}$ | 0.0052 | 0.0210 |
| Cross- entropy error | 0.0036 | 0.0134 | 0.0188 |
| Minkowski error | 0.0459 | 0.2236 | 0.4722 |
| Weighted squared error | $9.8919e^{-05}$ | 0.0122 | 0.0241 |

**Table 4:** All performance measures (Testing errors).

### 3.2 Confusion table

This section shows the correct and miss classifications made by the model on the testing dataset. The table 5 below is termed as a confusion matrix. The rows in the table denotes the target variables, whereas the columns represent the output classes belonging to the testing dataset. The diagonal cells depict the correctly classified, and the off-diagonal cells indicates the misclassified instances. The decision threshold is considered to be 0.5. The table shows 61 correctly classified instances and one misclassified instance.

| Instances | Predicted positive | Predicted negative |
|---|---|---|
| Actual positive | 33 | 0 |
| Actual negative | 1 | 28 |

**Table 5:** Confusion table showing positive and negative classification.

Classification accuracy was then calculated from the confusion matrix using the following equation 2:

[(Correct prediction) / (Total predicted)] × 100 (2)

The following part of this section will evaluate the model’s performance based on ROC, Cumulative gain, Lift plot and Loss index.

### 3.3 Receiver operating characteristic

A good way to analyze the loss is by plotting a ROC (Receiver Operating Characteristic) curve which is a graphical illustration of how good the classifier distinguishes between the two dissimilar classes. This capacity for discrimination is measured by calculating the area under the curve. ROC was found to be 1 with an optimal threshold of 0.882. The Figure 4 below shows the ROC curve obtained and the blue shaded region is the AUC.

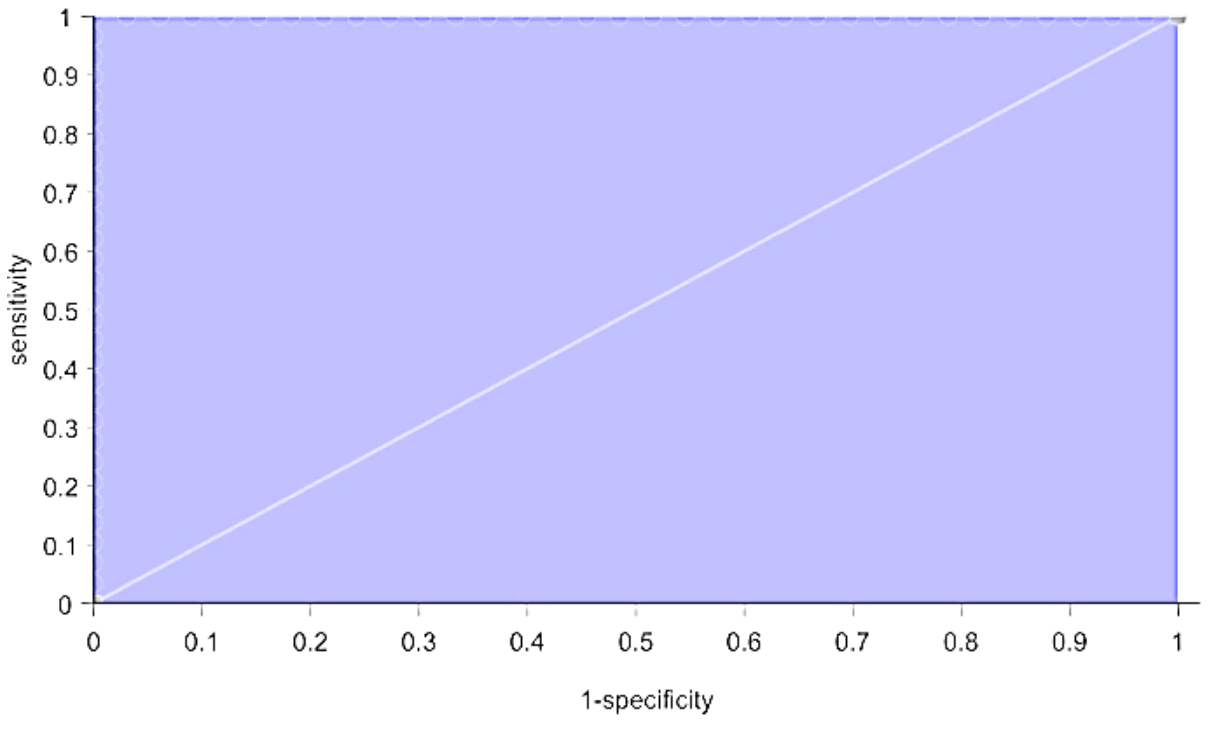

**Figure 4:** ROC plot.

### 3.4 Cumulative gain

The cumulative gain analysis is a pictorial representation that illustrates the benefit of implementing a predictive model as opposed to randomness. The baseline represents the results that would be obtained without using a model. The positive cumulative gain which shows on the y-axis the percentage of positive instances found against the percentage of the population, which is represented on the x-axis. Similarly, the negative cumulative gain represents the fraction of the negative instances found against the population. The following Figure 5 shows the results of the analysis in this case. The blue line represents the positive cumulative gain, the red line shows the negative cumulative gain, and the grey line signifies the random cumulative gain. Greater separation between the positive and negatives cumulative gain charts, indicates better performance of the classifier.

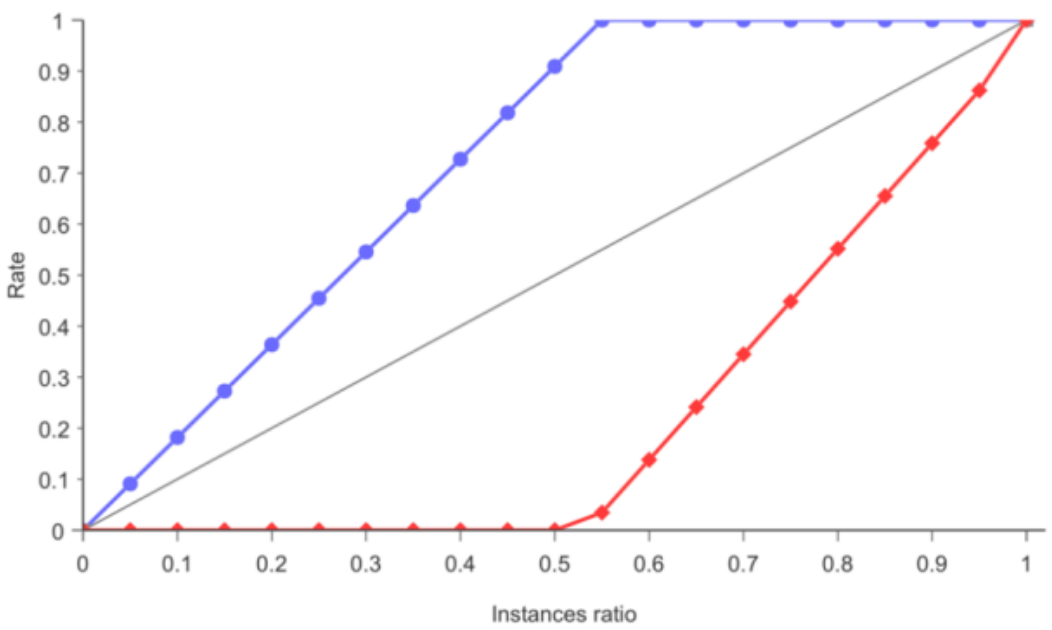

**Figure 5:** Cumulative plot.

The maximum gain score of a cumulative gain is obtained by the computing the maximum difference between the positive and the negative cumulative gain, i.e., it is the point where the percentage of positive instances found is maximized, and the percentage of negative instances found is minimized. If we have a perfect model, this score takes the value 1. The next table 6 shows the score in this case and the instances ratio for which it is reached.

| Parameters | Value |
|---|---|
| Instance ratio | 0.55 |
| Maximum gain score | 0.97 |

**Table 6:** Maximum gain score.

### 3.5 Lift plot

A lift plot provides a visual aid for evaluating a predictive model loss. It consists of a lift curve and a baseline. Lift curve represents the ratio between the positive events using a model and without using it. Baseline represents randomness, i.e., not using a model. The Figure 6 below shows the lift plot obtained for this study. The x-axis displays the percentage of instances studied while the y-axis displays the ratio between the results predicted by the model and the results without the model.

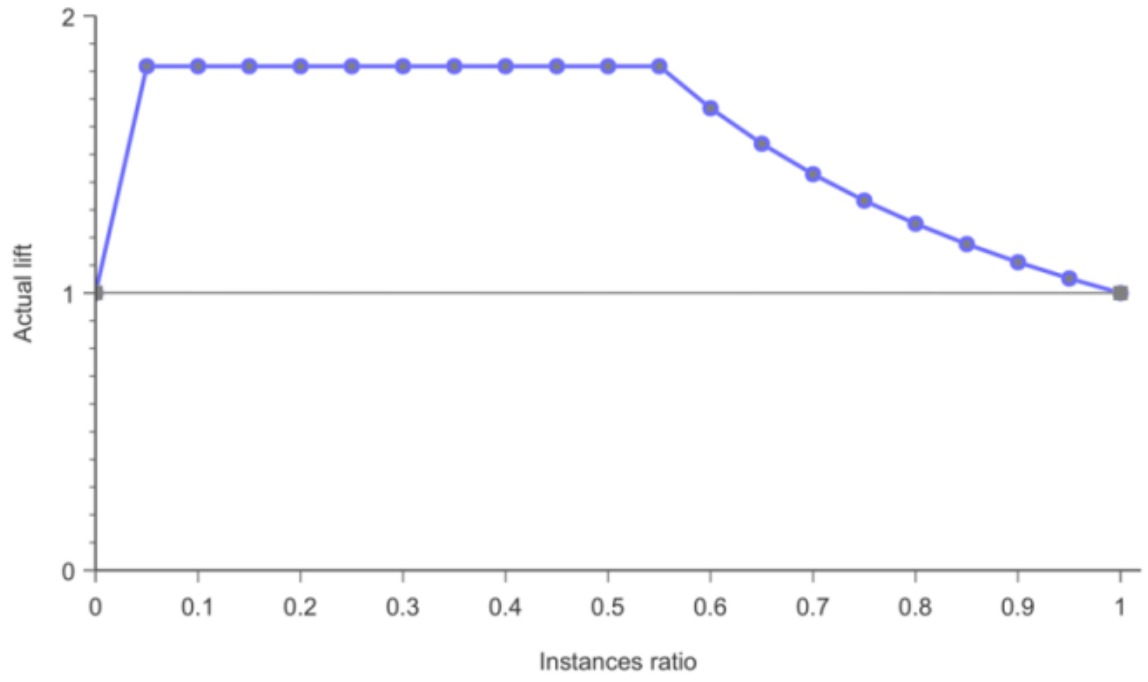

**Figure 6:** Lift plot.

### 3.6 Loss index

The loss index plays an essential role in the use of a neural network. It delineates the errand that the neural network is designed to perform and provides a gauge of the quality of the desired learning. The loss index varies with different applications. We implemented the weighted squared error as the error method which is especially useful when the data set has unbalanced targets. That is, there are too few positives when compared to the negatives, or vice versa. The following table 7 shows the exponent in error between the outputs from the neural network and the targets in the dataset.

| Parameters | Value |
|---|---|
| Positive weight | 2.72 |
| Negative weight | 1 |

**Table 7:** Exponent in error between the output from the ANN and the target variable.

## 4. Conclusion

To our knowledge, this is the first attempt to examine identification of autism using the Levenberg-Marquardt algorithm and incremental order selection in the field of artificial neural networks. Moreover, our model produced the highest classification accuracy of 98.38%. Through this study, we highlighted the importance of the training algorithm, order selection algorithm and selecting the required loss index to deal with unbalanced data. Based on the observations and evaluation of the proposed model, it can be inferred that Neural network with the Levenberg-Marquardt algorithm and incremental order selection is an appropriate tool for diagnosing ASD and can be deployed as a clinical decision support system.

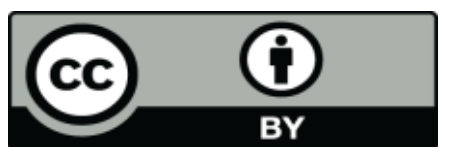